\documentclass{article}
\usepackage{graphicx}
\usepackage{booktabs}
\usepackage{multirow}
\usepackage[table]{xcolor} 
\usepackage{amssymb}
\usepackage{pifont}
\usepackage[font=small,skip=8pt]{caption}
\usepackage{wrapfig}
\usepackage{amsmath}

 \usepackage[sglblindworkshop, final]{neurips_2025}

\usepackage[utf8]{inputenc} 
\usepackage[T1]{fontenc}    
\usepackage{hyperref}       
\usepackage{url}            
\usepackage{booktabs}       
\usepackage{amsfonts}       
\usepackage{nicefrac}       
\usepackage{microtype}      
\usepackage{xcolor} 
\usepackage{adjustbox} 

\title{Dual Mixture-of-Experts Framework for Discrete-Time Survival Analysis}
\workshoptitle{Learning from Time Series for Health (TS4H)}

\author{%
  Hyeonjun Lee\\
  Lunit Inc.\\
  \texttt{hyeonjun1882@lunit.io} \\
  \And
  Hyungseob Shin\\
  Lunit Inc.\\
  \texttt{hs.shin@lunit.io} \\
  \And
  Gunhee Nam\\
  Lunit Inc.\\
  \texttt{ghnam@lunit.io} \\
  \And
  Hyeonsoo Lee\\
  Lunit Inc.\\
  \texttt{hslee@lunit.io}
}

\begin{document}

\maketitle

\begin{abstract}
Survival analysis is a task to model the time until an event of interest occurs, widely used in clinical and biomedical research. 
A key challenge is to model patient heterogeneity while also adapting risk predictions to both individual characteristics and temporal dynamics.
We propose a dual mixture-of-experts (MoE) framework for discrete-time survival analysis. Our approach combines a feature-encoder MoE for subgroup-aware representation learning with a hazard MoE that leverages patient features and time embeddings to capture temporal dynamics. This dual-MoE design flexibly integrates with existing deep learning–based survival pipelines. On METABRIC and GBSG breast cancer datasets, our method consistently improves performance, boosting the time-dependent C-index up to 0.04 on the test sets, and yields further gains when incorporated into the Consurv framework. 
\end{abstract}    
\section{Introduction}

Survival analysis aims to predict the time until an event while properly accounting for censoring.
A long-standing approach is the Cox Proportional Hazards (CPH) model~\citep{Cox1972JRSSB}, which assumes that hazard ratios between patients remain proportional over time. While effective in many settings, this assumption often fails in real-world clinical data, where risk dynamics are non-proportional over time. To address this limitation, recent deep learning models (e.g., DeepHit~\citep{lee2018deephit}, ConSurv~\citep{lee2024toward}) replace the CPH constraint with flexible neural architectures and are trained with negative log-likelihood objectives~\citep{Gensheimer2019PeerJ,Ren2019DRSA,curth2021survite,lee2024toward}, enabling the modeling of non-proportional hazards.

Despite this progress, most deep survival models still rely on a single shared feature encoder. In practice, patients form heterogeneous subgroups with distinct risk profiles, and a single encoder tends to favor dominant patterns while underrepresenting minority subgroups~\citep{Zhou_2021, guo-etal-2018-multi, Jin_2023_CVPR}. Hazard estimation is likewise commonly implemented with a single network, yet survival risk is both time-varying and patient-specific: two patients at the same time point may exhibit markedly different risk trajectories depending on their clinical characteristics. A single network implicitly ties all patients and all time bins to one shared functional form, leaving further room for improvement in capturing patient heterogeneity and temporal dynamics.

\begin{figure*}[t]
    \begin{center}
    \includegraphics[scale=0.28]
    {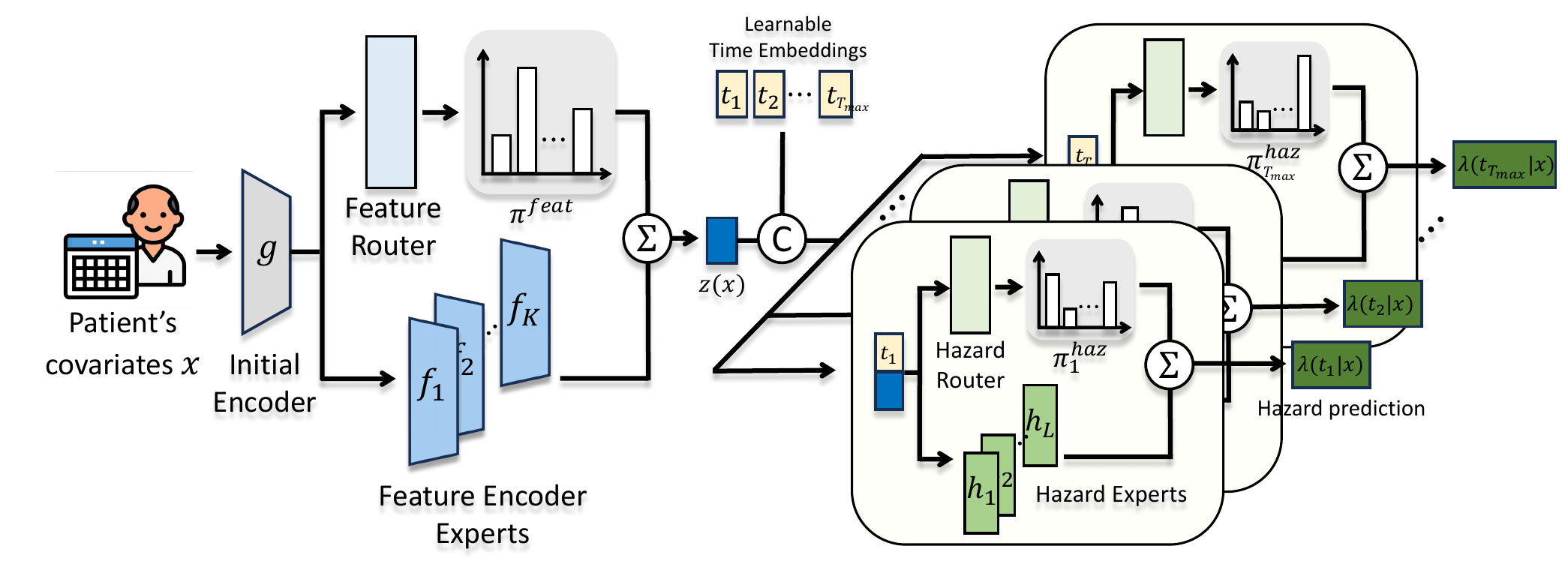}
    \captionsetup{labelformat=empty}
    \caption{\textbf{Figure 1.} Overall architecture of the proposed framework. Unlike prior survival models that use a single encoder and a single hazard head, our framework employs dual mixtures of experts: one over feature encoders and another over hazard networks, where hazard experts are shared across time bins and dynamically routed by patient and time embeddings.}
    \vspace{-0.6cm}
    \label{fig:method-overview}
    \end{center}
\end{figure*}

In this paper, we propose a dual mixture-of-experts (MoE)~\citep{shazeer2017outrageouslylargeneuralnetworks} framework that integrates a mixture of feature encoders and a mixture of hazard networks to address these limitations (Fig.~\ref{fig:method-overview}). The feature-encoder MoE models patient heterogeneity through soft routing based on each patient’s encoded features, while the hazard-network MoE outputs a full hazard vector over the prediction horizon with a soft router conditioned on both patient representation and time embedding. This joint design enables experts to specialize along temporal horizons while adapting to patient subgroups, resulting in finer-grained, context-aware hazard modeling. Experiments on the METABRIC and GBSG datasets show consistent improvements in both overall and time-dependent C-index over conventional single-network models, with further gains when incorporated into the ConSurv framework.

\section{Method}

\subsection{Preliminary}
We formulate survival prediction in discrete-time setting. Each patient $i \in \{1,...,N\}$ is represented by covariates $x_i$, an observed time $\tau_i \in \{0,...,T_{max}\}$, and an event indicater $\delta_i \in \{0,1\}$, where $\delta_i =1$ denotes an obeserved event at $\tau_i$ and $\delta_i = 0$ indicates that the observation is right-censored at $\tau_i$; i.e., the patient was event-free up to $\tau_i$ but their subsequent status is unobserved.

The conditional hazard function specifies the instantaneous event probability at time $t$:
\[
\lambda(t \mid x) = \mathbb{P}(T = t \mid T \geq t, x).
\]

It induces the survival function
\[
S(t \mid x) = \prod_{t' \leq t} \big(1 - \lambda(t' \mid x)\big),
\]
which captures the probability of remaining event-free after $t$. 
The probability mass for an event at time $t$ is then
\[
p(t \mid x) = \lambda(t \mid x)\, S(t-1 \mid x).
\]
Model estimation proceeds via maximum likelihood. 
The negative log-likelihood objective combines information from observed and censored cases:
\[
\mathcal{L}_{\text{NLL}} = 
- \sum_{i=1}^{N} \Big[ 
\delta_i \log \hat{p}(\tau_i \mid x_i) 
+ (1 - \delta_i) \log \hat{S}(\tau_i \mid x_i) 
\Big].
\]
\subsection{Mixture of Feature Encoders}
We first enhance the representation learning stage by introducing a mixture of feature encoder architecture.
An initial encoder $g(\cdot)$ extracts patient-level representations, which are then routed into multiple expert encoders $\{f_k\}_{k=1}^K$. 
The router takes patient features as input and produces $\pi^{feat}_k$ via softmax function, which represents the routing probability (or mixing weight) of expert $k$ for patient $x$. 
The final encoded representation for patient $x$ is computed as
\[
z(x) = \sum_{k=1}^{K} \pi^{feat}_k \cdot f_k(g(x))
\]
By conditioning routing decisions on patient features, this design encourages the encoder to discover hidden subgroups and produce subgroup-aware representations. 
To prevent collapse into a single expert, we incorporate a load balancing loss $\mathcal{L}^{\text{feat}}_{LB} = \alpha \Big( K \sum_k \bar{\pi}_k^{\text{feat}^2} - 1 \Big)$
, where $\bar{\pi}_k^{\text{feat}} = \frac{1}{B} \sum_{i=1}^B \pi^{feat}_{k,i}$ denotes the batch wise averaged assignment probability for routing logit for expert $k$.
This regularizer promotes healthy utilization of all experts by penalizing excessive reliance on a subset of them. 

\subsection{Mixture of Hazard Networks}
On top of the encoded patient representations, we further introduce a mixture of hazard network for hazard prediction. 
Unlike feature encoder MoE, the hazard MoE conditions routing on both patient features and temporal embeddings.
Each hazard expert $\{h_l\}_{l=1}^L$ predicts hazards for all discrete time bins, while the router conditions by concatenation of patient features and time embeddings to produce routing probability $\pi^{haz}_{t,l}$. 
The final hazard prediction at $t$ is expressed as
\[
\lambda(t|x) = \sum_{l=1}^{L} \pi^{haz}_{t,l} \cdot h_l(z(x), e_t)
\]
where $e_t$ denotes the learnable time embedding for time-bin $t$.
Joint conditioning on patient features and time enables experts to specialize across both patient heterogeneity and temporal dynamics. This enables experts to capture finer-grained survival patterns (e.g., subgroups that differ not only in patient profiles but also in how risks evolve over time). 
As with the feature encoder MoE, we apply a load balancing loss $\mathcal{L}_{LB}^{haz} = \beta \big( T_{max} \sum_t  L \sum_l {\bar{\pi}_{t,l}^{\text{haz}^2}} - 1 \big)$ to encourage balanced usage of hazard experts across all time-bin.

\subsection{Overall Training Objective}
The overall training objective combines the discrete-time negative log-likelihood (NLL) loss with the load balancing regularizers applied at both stages:
\[
\mathcal{L} = \mathcal{L}_{NLL} +  \mathcal{L}_{LB}^{feat} +  \mathcal{L}_{LB}^{haz}.
\]
This formulation ensures that the model not only fits observed survival outcomes but also maintains balanced expert utilization across both representation and prediction stages, leading to more robust subgroup- and time-aware survival modeling.

\section{Experiment}
\newcommand{\pmval}[2]{#1 {\scriptsize $\pm$ #2}}

\begin{table}[h!]
\centering
\resizebox{\textwidth}{!}{%
\begin{tabular}{c c c *{9}{c}}
\toprule
\multirow{2}{*}{\textbf{Method}} & 
\multirow{2}{*}{\textbf{Dual MoE}} & 
\multirow{2}{*}{\textbf{C-index}} & 
\multicolumn{9}{c}{\textbf{Time-dependent C-index}} \\
\cmidrule(lr){4-12}
 & & & 10\% & 20\% & 30\% & 40\% & 50\% & 60\% & 70\% & 80\% & 90\% \\
\midrule
\multicolumn{12}{c}{\cellcolor{gray!20}\textit{Metabric}} \\  
\midrule
CoxPH~\cite{Cox1972JRSSB} & - & \pmval{0.663}{0.017}
 & \pmval{0.658}{0.032} 
 & \pmval{0.667}{0.022}
 & \pmval{0.665}{0.018} 
 & \pmval{0.665}{0.019}
 & \pmval{0.659}{0.013} 
 & \pmval{0.651}{0.015}
 & \pmval{0.648}{0.015}
 & \pmval{0.639}{0.020} 
 & \pmval{0.646}{0.026} \\
 \midrule
Naïve impl. & \ding{55} & \pmval{0.646}{0.021} & \cellcolor{yellow!25}\pmval{0.670}{0.050} & \pmval{0.660}{0.032} & \pmval{0.644}{0.019} & \pmval{0.644}{0.021} & \pmval{0.638}{0.022} & \pmval{0.629}{0.015} & \pmval{0.621}{0.017} & \pmval{0.611}{0.024} & \pmval{0.606}{0.022}
  \\
Naïve impl. & \ding{51} &  \cellcolor{yellow!25}\pmval{0.654}{0.015} & \pmval{0.669}{0.032} &  \cellcolor{yellow!25}\pmval{0.667}{0.022} &  \cellcolor{yellow!25}\pmval{0.657}{0.013} &  \cellcolor{yellow!25}\pmval{0.653}{0.017} &  \cellcolor{yellow!25}\pmval{0.646}{0.015} &  \cellcolor{yellow!25}\pmval{0.638}{0.016} &  \cellcolor{yellow!25}\pmval{0.628}{0.018} &  \cellcolor{yellow!25}\pmval{0.621}{0.027} & \ \cellcolor{yellow!25}\pmval{0.623}{0.022}
\\
\midrule
ConSurv~\cite{lee2024toward} & \ding{55} & \pmval{0.657}{0.020} & \pmval{0.656}{0.044} & \pmval{0.668}{0.030} & \pmval{0.658}{0.018} & \pmval{0.657}{0.021} & \pmval{0.649}{0.018} & \pmval{0.639}{0.011} & \pmval{0.629}{0.010} & \pmval{0.616}{0.024} & \pmval{0.617}{0.026}
 \\
ConSurv~\cite{lee2024toward} & \ding{51} & \cellcolor{yellow!25}\pmval{0.668}{0.018} &  \cellcolor{yellow!25}\pmval{0.696}{0.034} & 
 \cellcolor{yellow!25}\pmval{0.689}{0.024} & 
  \cellcolor{yellow!25}\pmval{0.676}{0.021} & 
   \cellcolor{yellow!25}\pmval{0.669}{0.022} & 
    \cellcolor{yellow!25}\pmval{0.657}{0.017} & 
     \cellcolor{yellow!25}\pmval{0.647}{0.015} & 
      \cellcolor{yellow!25}\pmval{0.642}{0.016} & 
       \cellcolor{yellow!25}\pmval{0.632}{0.021} & 
        \cellcolor{yellow!25}\pmval{0.634}{0.019}
 \\ 
 \midrule
\multicolumn{12}{c}{\cellcolor{gray!20}\textit{GBSG}} \\ 
\midrule
CoxPH~\cite{Cox1972JRSSB} & - & \pmval{0.659}{0.012}  & \pmval{0.739}{0.046} & \pmval{0.709}{0.018} &\pmval{0.681}{0.017}  & \pmval{0.676}{0.014} & \pmval{0.670}{0.013} & \pmval{0.662}{0.012} & \pmval{0.658}{0.011} & \pmval{0.655}{0.011} & \pmval{0.652}{0.011}   \\
\midrule
Naïve impl. & \ding{55} & \pmval{0.662}{0.012} & \pmval{0.744}{0.039} & \pmval{0.706}{0.017} & \pmval{0.678}{0.018} &
\pmval{0.674}{0.015} & \pmval{0.669}{0.014} & \pmval{0.662}{0.012} &
\pmval{0.657}{0.011} & \pmval{0.655}{0.012} & \pmval{0.652}{0.011}  \\
Naïve impl. & \ding{51} & \cellcolor{yellow!25}\pmval{0.667}{0.010} & \cellcolor{yellow!25}\pmval{0.751}{0.033} & \cellcolor{yellow!25}\pmval{0.717}{0.018} & \cellcolor{yellow!25}\pmval{0.689}{0.016} &
\cellcolor{yellow!25}\pmval{0.684}{0.016} & \cellcolor{yellow!25}\pmval{0.677}{0.014} & \cellcolor{yellow!25}\pmval{0.670}{0.011} &
\cellcolor{yellow!25}\pmval{0.666}{0.011} & \cellcolor{yellow!25}\pmval{0.663}{0.011} & \cellcolor{yellow!25}\pmval{0.659}{0.010} \\
\midrule
ConSurv~\cite{lee2024toward} & \ding{55} & \pmval{0.665}{0.011} & \pmval{0.742}{0.039} & \pmval{0.709}{0.017} & \pmval{0.682}{0.016} &
\pmval{0.679}{0.014} & \pmval{0.674}{0.013} & \pmval{0.667}{0.011} &
\pmval{0.663}{0.011} & \pmval{0.661}{0.011} & \pmval{0.658}{0.010} \\
ConSurv~\cite{lee2024toward} & \ding{51} & \cellcolor{yellow!25}\pmval{0.668}{0.011} & \cellcolor{yellow!25}\pmval{0.752}{0.036} & \cellcolor{yellow!25}\pmval{0.715}{0.018} & \cellcolor{yellow!25}\pmval{0.689}{0.017} &
\cellcolor{yellow!25}\pmval{0.684}{0.016} & \cellcolor{yellow!25}\pmval{0.677}{0.014} & \cellcolor{yellow!25}\pmval{0.670}{0.012} &
\cellcolor{yellow!25}\pmval{0.666}{0.011} & \cellcolor{yellow!25}\pmval{0.663}{0.011} & \cellcolor{yellow!25}\pmval{0.659}{0.010}\\

\bottomrule
\end{tabular}
}
\captionsetup{labelformat=empty}
\vspace{1mm}
\caption{\textbf{Table 1.} Comparison of both overall and time-dependent C-index on the Metabric and GBSG datasets. We report the average performance over 10 random seeds.\protect\footnotemark}
\vspace{-6mm}
\label{tab:main}
\end{table}

\subsection{Experimental Settings}
\textbf{Dataset.} We evaluate our method on two widely used breast cancer survival datasets, which are Metabric~\citep{Curtis2012Nature} and GBSG~\citep{Schumacher1994JCO} . Metabric contains clinical and gene expression information from 1,981 patients, with 21 variables in total; 55.2\% of the cases are censored and 44.8\% uncensored. GBSG includes 2,232 patients with 21 clinical and tumor-related variables, originally collected to study the impact of hormone therapy on recurrence-free survival. In this dataset, 43.2\% of the cases are censored and 56.8\% uncensored.

\footnotetext{Results may differ from ~\citep{lee2024toward} since dataset splits vary with different random seeds.}
\textbf{Comparison Methods.} We evaluated the effectiveness of the proposed dual mixture-of-experts framework under two experimental settings. First, we considered a naïve implementation trained solely with $\mathcal{L}_{NLL}$, consisting of a single feature encoder and a single hazard network, and compared it against our proposed dual MoE framework. Second, we applied our approach on top of the ConSurv framework~\citep{lee2024toward}, replacing its feature encoder and hazard network with a mixture of feature encoders and a mixture of hazard networks, respectively. In both cases, we set the number of feature-encoder experts ($K$) and hazard experts ($L$) to (4, 4) for Metabric and (6, 3) for GBSG. More details can be found in the Appendix.

\textbf{Evaluation Metrics.} 
We evaluate performance using the concordance index (C-index)~\citep{Harrell1984StatMed}, which measures concordance between predicted hazards and observed event times across the entire period. 
However, this global measure may overlook how performance varies across different time intervals. 
To assess temporal variations, we also report time-dependent C-index, computed at multiple time horizons defined by the 10\%–90\% percentiles of observed event times~\citep{Gerds2013StatMed}.


\subsection{Results}
Table~\ref{tab:main} summarizes the main performance comparison of the proposed dual MoE framework.
By replacing the single feature encoder and hazard network with dual mixtures, we observed consistent improvements in both overall and time-dependent C-index on the METABRIC and GBSG datasets.
Furthermore, integrating our framework with ConSurv leads to additional gains, indicating that the proposed method is easily applicable to other deep learning based discrete-time survival models.

\begin{figure}[t]
\hspace{10mm} 
\scalebox{0.82}{
\begin{minipage}{0.38\linewidth}
\includegraphics[width=\linewidth]{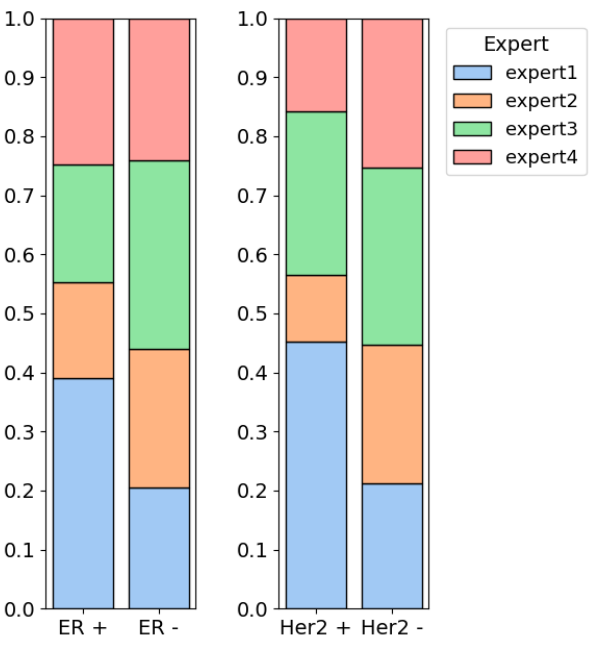}
\captionsetup{labelformat=empty}
\captionof{figure}{\textbf{Figure 2.} Average routing probabilities of \textit{feature-encoder experts} across ER and HER2 subgroups. }
\label{fig:feature-router-subgroup}
\end{minipage}\hspace{20mm}\hfill
\begin{minipage}{0.55\linewidth}
\centering 
\includegraphics[width=\linewidth]{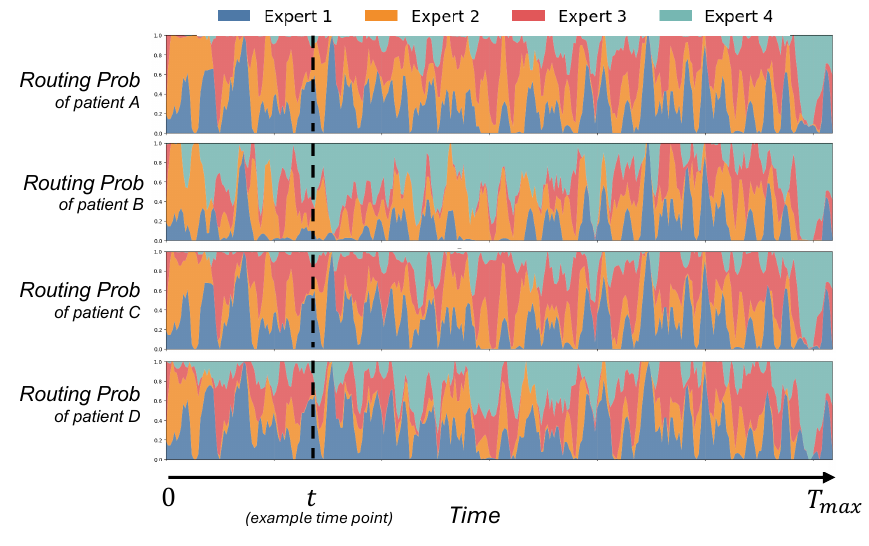}
\captionsetup{labelformat=empty}
\captionof{figure}{\textbf{Figure 3.} Patient-level routing probabilities of \textit{hazard experts} over time. 
At any discrete time bin $t$ (a vertical slice), the proportion of each color equals proportion of routing probability to each hazard experts ($\pi^{haz}_{t,l}, l \in \{1,2,3,4\}$), which sum to $1$.
}
\label{fig:haard-router-patient}
\end{minipage}
}
\end{figure}


\textbf{Visualization of Feature Routing Probability.}
We examined the routing behavior of the feature-encoder MoE across estrogen receptor (ER) and HER2 subgroups using Metabric dataset (Fig.~\ref{fig:feature-router-subgroup}). Specifically, for each subgroup, we averaged the feature-encoder routing probabilities across all patients belonging to that subgroup to obtain representative expert assignment distributions. The router exhibited distinct expert preferences between subgroups, indicating that it adapts to patient heterogeneity rather than assigning weights uniformly.

\textbf{Visualization of Trajectory of Hazard Routing Probability through Time.}
We visualized patient-level routing probabilities of the hazard router (Fig.~\ref{fig:haard-router-patient}). Each panel shows the soft assignment probabilities of four hazard experts over time for each of 4 patients (A-D). Routing patterns vary across patients but consistently show shifts in expert dominance between early and late time horizons, indicating adaptation to both individual heterogeneity and temporal structure.

More ablation studies, including effect of each MoE architecture and input of hazard router, are provided in the Appendix.

\section{Conclusion \& Future Work}
In this work, we proposed a dual mixture-of-experts framework for discrete-time survival analysis that integrates mixtures of feature encoders and hazard networks to model patient heterogeneity and temporal risk variation. On the Metabric and GBSG datasets, our method consistently outperformed conventional single-network models and yielded further gains with ConSurv. As future work, we plan to further analyze the role of each expert and extend our framework to multimodal settings, such as mammography-based risk prediction.
{
    \small
    \bibliographystyle{plainnat}
    \bibliography{main}

\begin{thebibliography}{14}
\providecommand{\natexlab}[1]{#1}
\providecommand{\url}[1]{\texttt{#1}}
\expandafter\ifx\csname urlstyle\endcsname\relax
  \providecommand{\doi}[1]{doi: #1}\else
  \providecommand{\doi}{doi: \begingroup \urlstyle{rm}\Url}\fi

\bibitem[Cox(1972)]{Cox1972JRSSB}
David~R. Cox.
\newblock Regression models and life-tables.
\newblock \emph{Journal of the Royal Statistical Society: Series B (Methodological)}, 34\penalty0 (2):\penalty0 187--202, 1972.
\newblock \doi{10.1111/j.2517-6161.1972.tb00899.x}.
\newblock URL \url{https://doi.org/10.1111/j.2517-6161.1972.tb00899.x}.

\bibitem[Curth et~al.(2021)Curth, Lee, and van~der Schaar]{curth2021survite}
Alicia Curth, Changhee Lee, and Mihaela van~der Schaar.
\newblock Surv{ITE}: Learning heterogeneous treatment effects from time-to-event data.
\newblock In A.~Beygelzimer, Y.~Dauphin, P.~Liang, and J.~Wortman Vaughan, editors, \emph{Advances in Neural Information Processing Systems}, 2021.
\newblock URL \url{https://openreview.net/forum?id=f0_tkoEJV88}.

\bibitem[Curtis et~al.(2012)Curtis, Shah, Chin, Turashvili, Rueda, Dunning, Speed, Lynch, Samarajiwa, Yuan, Gr{\"a}f, Ha, Haffari, Bashashati, Russell, McKinney, Group, Langer{\o}d, Green, Provenzano, Wishart, Pinder, Watson, Markowetz, Murphy, Ellis, Purushotham, B{\o}rresen{-}Dale, Brenton, Tavar{\'e}, Caldas, and Aparicio]{Curtis2012Nature}
Christina Curtis, Sohrab~P. Shah, Suet-Feung Chin, Gulisa Turashvili, Oscar~M. Rueda, Mark~J. Dunning, Doug Speed, Andy~G. Lynch, Shamith Samarajiwa, Yinyin Yuan, Stefan Gr{\"a}f, Gavin Ha, Gholamreza Haffari, Ali Bashashati, Roslin Russell, Steven McKinney, METABRIC Group, Anita Langer{\o}d, Andrew Green, Elena Provenzano, Gordon Wishart, Sarah Pinder, Peter Watson, Florian Markowetz, Leigh Murphy, Ian Ellis, Arnie Purushotham, Anne{-}Lise B{\o}rresen{-}Dale, James~D. Brenton, Simon Tavar{\'e}, Carlos Caldas, and Samuel Aparicio.
\newblock The genomic and transcriptomic architecture of 2,000 breast tumours reveals novel subgroups.
\newblock \emph{Nature}, 486\penalty0 (7403):\penalty0 346--352, 2012.
\newblock \doi{10.1038/nature10983}.
\newblock URL \url{https://doi.org/10.1038/nature10983}.
\newblock PMCID: PMC3440846.

\bibitem[Gensheimer and Narasimhan(2019)]{Gensheimer2019PeerJ}
Michael~F. Gensheimer and Balasubramanian Narasimhan.
\newblock {A scalable discrete-time survival model for neural networks}.
\newblock \emph{PeerJ}, 7:\penalty0 e6257, 2019.
\newblock \doi{10.7717/peerj.6257}.
\newblock URL \url{https://doi.org/10.7717/peerj.6257}.

\bibitem[Gerds et~al.(2013)Gerds, Kattan, Schumacher, and Yu]{Gerds2013StatMed}
Thomas~A. Gerds, Michael~W. Kattan, Martin Schumacher, and Chang Yu.
\newblock Estimating a time-dependent concordance index for survival prediction models with covariate dependent censoring.
\newblock \emph{Statistics in Medicine}, 32\penalty0 (13):\penalty0 2173--2184, June 2013.
\newblock \doi{10.1002/sim.5681}.
\newblock Epub 2012 Nov 22.

\bibitem[Guo et~al.(2018)Guo, Shah, and Barzilay]{guo-etal-2018-multi}
Jiang Guo, Darsh Shah, and Regina Barzilay.
\newblock Multi-source domain adaptation with mixture of experts.
\newblock In Ellen Riloff, David Chiang, Julia Hockenmaier, and Jun{'}ichi Tsujii, editors, \emph{Proceedings of the 2018 Conference on Empirical Methods in Natural Language Processing}, pages 4694--4703, Brussels, Belgium, October-November 2018. Association for Computational Linguistics.
\newblock \doi{10.18653/v1/D18-1498}.
\newblock URL \url{https://aclanthology.org/D18-1498/}.

\bibitem[Harrell et~al.(1984)Harrell, Lee, Califf, Pryor, and Rosati]{Harrell1984StatMed}
Frank E.~Jr Harrell, Kerry~L. Lee, Robert~M. Califf, David~B. Pryor, and Robert~A. Rosati.
\newblock Regression modelling strategies for improved prognostic prediction.
\newblock \emph{Statistics in Medicine}, 3\penalty0 (2):\penalty0 143--152, April--June 1984.
\newblock \doi{10.1002/sim.4780030207}.

\bibitem[Jin et~al.(2023)Jin, Li, Lu, Cheung, and Wang]{Jin_2023_CVPR}
Yan Jin, Mengke Li, Yang Lu, Yiu-ming Cheung, and Hanzi Wang.
\newblock Long-tailed visual recognition via self-heterogeneous integration with knowledge excavation.
\newblock In \emph{Proceedings of the IEEE/CVF Conference on Computer Vision and Pattern Recognition (CVPR)}, pages 23695--23704, June 2023.

\bibitem[Lee et~al.(2018)Lee, Zame, Yoon, and van~der Schaar]{lee2018deephit}
Changhee Lee, William Zame, Jinsung Yoon, and Mihaela van~der Schaar.
\newblock Deephit: A deep learning approach to survival analysis with competing risks.
\newblock In \emph{Proceedings of the AAAI Conference on Artificial Intelligence}, volume~32, 2018.
\newblock \doi{10.1609/aaai.v32i1.11842}.
\newblock URL \url{https://doi.org/10.1609/aaai.v32i1.11842}.

\bibitem[Lee et~al.(2024)Lee, Park, and Lee]{lee2024toward}
Dongjoon Lee, Hyeryn Park, and Changhee Lee.
\newblock Toward a well-calibrated discrimination via survival outcome-aware contrastive learning.
\newblock In \emph{The Thirty-eighth Annual Conference on Neural Information Processing Systems}, 2024.
\newblock URL \url{https://openreview.net/forum?id=UVjuYBSbCN}.

\bibitem[Ren et~al.(2019)Ren, Qin, Zheng, Yang, Zhang, Qiu, and Yu]{Ren2019DRSA}
Kan Ren, Jiarui Qin, Lei Zheng, Zhengyu Yang, Weinan Zhang, Lin Qiu, and Yong Yu.
\newblock {Deep Recurrent Survival Analysis}.
\newblock In \emph{Proceedings of the Thirty-Third AAAI Conference on Artificial Intelligence (AAAI-19)}, pages 4798--4805. AAAI Press, 2019.
\newblock ISBN 978-1-57735-809-1.
\newblock \doi{10.1609/aaai.v33i01.33014798}.
\newblock URL \url{https://doi.org/10.1609/aaai.v33i01.33014798}.

\bibitem[Schumacher et~al.(1994)Schumacher, Bastert, Bojar, H{\"u}bner, Olschewski, Sauerbrei, Schmoor, Beyerle, Neumann, and Rauschecker]{Schumacher1994JCO}
M.~Schumacher, G.~Bastert, H.~Bojar, K.~H{\"u}bner, M.~Olschewski, W.~Sauerbrei, C.~Schmoor, C.~Beyerle, R.~L. Neumann, and H.~F. Rauschecker.
\newblock Randomized 2 x 2 trial evaluating hormonal treatment and the duration of chemotherapy in node-positive breast cancer patients.
\newblock \emph{Journal of Clinical Oncology}, 12\penalty0 (10):\penalty0 2086--2093, October 1994.
\newblock \doi{10.1200/JCO.1994.12.10.2086}.

\bibitem[Shazeer et~al.(2017)Shazeer, Mirhoseini, Maziarz, Davis, Le, Hinton, and Dean]{shazeer2017outrageouslylargeneuralnetworks}
Noam Shazeer, Azalia Mirhoseini, Krzysztof Maziarz, Andy Davis, Quoc Le, Geoffrey Hinton, and Jeff Dean.
\newblock Outrageously large neural networks: The sparsely-gated mixture-of-experts layer, 2017.
\newblock URL \url{https://arxiv.org/abs/1701.06538}.

\bibitem[Zhou et~al.(2021)Zhou, Yang, Qiao, and Xiang]{Zhou_2021}
Kaiyang Zhou, Yongxin Yang, Yu~Qiao, and Tao Xiang.
\newblock Domain adaptive ensemble learning.
\newblock \emph{IEEE Transactions on Image Processing}, 30:\penalty0 8008–8018, 2021.
\newblock ISSN 1941-0042.
\newblock \doi{10.1109/tip.2021.3112012}.
\newblock URL \url{http://dx.doi.org/10.1109/TIP.2021.3112012}.

\end{thebibliography}
}

\appendix

\section{Training Details}

Table~\ref{tab:hyperparams_dataset} summarizes the key hyperparameters used in our experiments. 

\begin{table}[h]
\centering
\small
\begin{tabular}{lcc}
\toprule
\textbf{Item} & \textbf{METABRIC} & \textbf{GBSG} \\
\midrule 
Initial encoder & MLP (depth=4) & MLP (depth=3)  \\
Feature-encoder router & MLP (depth=1) & MLP (depth=1) \\
Feature-encoder expert & MLP (depth=1) & MLP (depth=2) \\
Number of feature-encoder experts ($K$) & 4 & 6 \\
Hazard router & MLP (depth=1) & MLP (depth=1) \\
Hazard expert & MLP (depth=1) & MLP (depth=1) \\
Number of hazard experts ($L$) & 4 & 3 \\
Time embedding dim ($d_{\text{time}}$) & 8 & 8 \\
Load-balancing coef. (feature) $\alpha$ & 0.3 & 0.3 \\
Load-balancing coef. (hazard) $\beta$ & 0.5 & 0.5 \\
\bottomrule
\end{tabular}
\vspace{3mm}
\captionsetup{labelformat=empty}
\captionof{table}{\textbf{Table 2.} Hyperparameter details. Here, depth refers to the number of hidden layers in each MLP.}
\label{tab:hyperparams_dataset}
\end{table}

\section{More Ablation Studies}

\begin{figure}[h]
\centering
\begin{minipage}{0.45\linewidth}
\centering
\footnotesize
\begin{tabular}{c c c}
\toprule
\textbf{Feat. MoE} & \textbf{Haz. MoE} & \textbf{C-index} \\
\midrule
\ding{55} & \ding{55} & \pmval{0.646}{0.021} \\
\ding{51} & \ding{55} & \pmval{0.649}{0.023} \\
\ding{55} & \ding{51} & \pmval{0.650}{0.025} \\
\ding{51} & \ding{51} & \cellcolor{yellow!25}\pmval{0.654}{0.015} \\
\bottomrule
\end{tabular}
\captionsetup{labelformat=empty}
\captionof{table}{\textbf{Table 3.} Ablation on each MoE architecture.}
\label{tab:ablation_moe}
\end{minipage}\hfill
\begin{minipage}{0.50\linewidth}
\centering
\includegraphics[width=\linewidth]{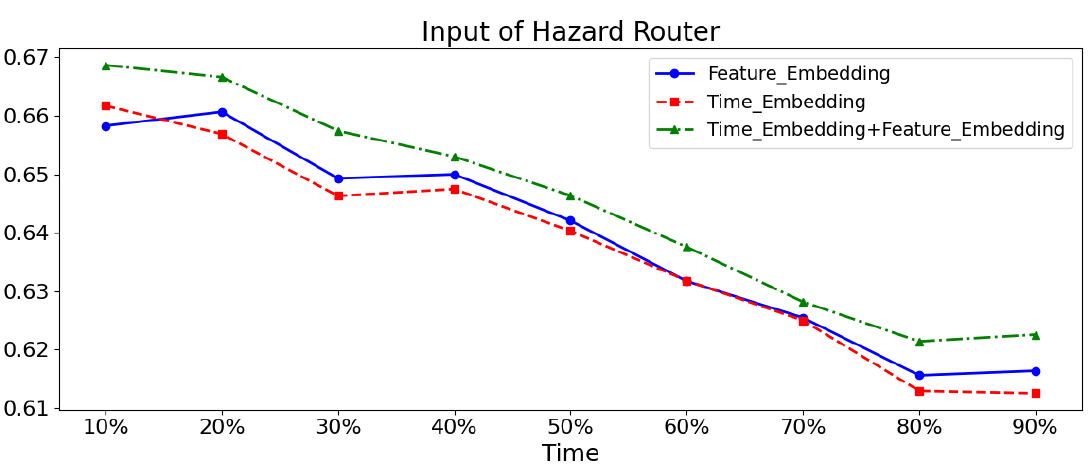}
\captionsetup{labelformat=empty}
\captionof{figure}
{\textbf{Figure 4.} Ablation on input of hazard router.}
\label{fig:hazard_input}
\end{minipage}
\end{figure}

For simplicity, we restrict our ablation experiments using Naïve implementation  using the Metabric dataset. 

\textbf{Effect of each MoE Architecture.}
We conducted ablation studies to evaluate the contribution of each MoE component, the mixture of feature encoders and the mixture of hazard networks.
As shown in Table~\ref{tab:ablation_moe}, introducing the mixture of feature encoders or the mixture of hazard networks individually improves performance over the conventional single-network models
When combined, the full dual mixture achieves the best performance, demonstrating that the two components are complementary.

\textbf{Input of Hazard Router.}
We evaluate how the choice of router inputs affects performance. Specifically, we measured time-dependent C-index for three variants: patient features only, time embeddings only, and both. Note that in all variants the hazard experts themselves still take both patient features and time embeddings as input; only the inputs to the router are modified.
As shown in Figure~\ref{fig:hazard_input}, using either patient features or time embeddings alone is suboptimal; their combination consistently achieves the best performance, highlighting the importance of jointly conditioning the router on both patient heterogeneity and temporal variation.

\end{document}